\documentclass[journal,10pt]{IEEEtran}
\usepackage{amsthm}
\usepackage{amsmath}
\usepackage{graphicx}
\usepackage{algorithm}
\usepackage{algorithmicx}
\usepackage{algpseudocode}
\usepackage{subfigure}
\usepackage{overpic}
\usepackage{colortbl}
\usepackage{pgfplots,tkz-fct}
\pgfplotsset{width=10cm, compat=1.16}

\ifCLASSINFOpdf
\else
\fi
\begin{document}
%
\title{PLMM: Personal Large Language Models on Mobile Devices}
%
%
%

\author{Yuanhao~Gong\\College of Electronics and Information Engineering, Shenzhen University, China.~~gong.ai@qq.com
	\thanks{Manuscript received April 19, 2005; revised September 17, 2014.}
	}

%
%

\markboth{Journal of \LaTeX\ Class Files,~Vol.~14, No.~8, August~2015}%
{Yuanhao: LLM on Blockchains}
%



\maketitle

\begin{abstract}
Inspired by Federated Learning, in this paper, we propose personal large models that are distilled from traditional large language models but more adaptive to local users' personal information such as education background and hobbies. We classify the large language models into three levels: the personal level, expert level and traditional level. The personal level models are adaptive to users' personal information. They encrypt the users' input and protect their privacy. The expert level models focus on merging specific knowledge such as finance, IT and art. The traditional models focus on the universal knowledge discovery and upgrading the expert models. In such classifications, the personal models directly interact with the user. For the whole system, the personal models have users' (encrypted) personal information. Moreover, such models must be small enough to be performed on personal computers or mobile devices. Finally, they also have to response in real-time for better user experience and produce high quality results. The proposed personal large models can be applied in a wide range of applications such as language and vision tasks.  
\end{abstract}


%
\IEEEpeerreviewmaketitle

%
%
%
%
\section{Introduction}
In recent years, the field of artificial intelligence has undergone a significant transformation thanks to the emergence of large language models, such as GPT-3~\cite{Brown2020}, T5~\cite{Raffel2020} and BERT~\cite{Devlin2019}. These models are powered by sophisticated algorithms and vast amounts of data, and they have revolutionized the way machines process and generate language. They can generate text that is remarkably similar to human-generated text, understand the context in which the text is being used, and even engage in conversations that are nearly indistinguishable from those between two humans.

The development of large language models is a result of deep learning techniques, which involve training on massive amounts of text data from the internet. These techniques enable the models to learn from the statistical patterns and linguistic features of the data, resulting in coherent and contextually relevant text. By developing models that can generate text that is remarkably similar to human-generated text, understand the context in which the text is being used, and even engage in conversations that are nearly indistinguishable from those between two humans, we can improve human-machine interaction.

Large language models have shown impressive capabilities across various tasks, including language translation, content creation, and chatbot development. They have shown great potential in numerous real-world applications, and their ability to understand, reason, and respond to a wide range of queries has made them invaluable tools for information retrieval and natural language processing tasks.

\begin{figure}
	\begin{minipage}{0.3\linewidth}
		\mbox{Personal level}
		\vfil
		\vspace{12mm}
		\mbox{Expert level}
		\vfil
		\vspace{4mm}
		\mbox{Traditional level}
	\end{minipage}
		\begin{minipage}{0.6\linewidth}
	\centering
	\includegraphics[width=\linewidth]{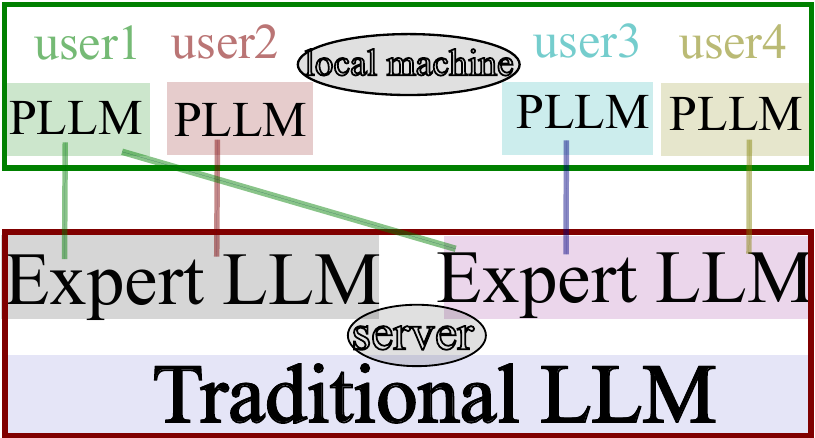}
\end{minipage}
\caption{The proposed personal large models run on local machines such as mobile phones and personal computers. Their parameters are trained by the user's personal input and also updated by the expert level models. The expert and traditional level large language models run on servers or cloud. The expert level models are good at specific fields, such as medical image analysis, finance, python coding, etc. The traditional models are good at merging the knowledge from all expert models and updating the expert models. This architecture is abbreviated as P-E-T for convenience and each level model is called P-model, E-model and T-model, respectively. }
\label{fig}
\end{figure}

One primary goal of developing large language models is to improve human-machine interaction. These models can help facilitate seamless communication and understanding between humans and machines by comprehending complex queries, providing insightful answers, and engaging in meaningful conversations. This has the potential to blur the lines between human and artificial intelligence and enhance the overall user experience.

However, the development of large language models also raises ethical concerns. Issues surrounding biases in the training data, potential misuse for spreading misinformation or generating harmful content, and privacy concerns have sparked debates within the AI community. To address these concerns and ensure the responsible deployment of large language models, proactive measures such as careful data selection, transparency, and accountability are necessary.

Despite these challenges, the advancements in large language models are progressing quickly. Researchers are continually working on improving the training methods and fine-tuning the models to enhance their performance. As these models become more accessible and widely used, we can expect to see further integration into various industries, powering innovations in healthcare, education, customer service, and beyond.

Large language models have revolutionized the field of artificial intelligence and opened up exciting possibilities for human-machine interaction. Their ability to understand and generate natural language has enormous potential for innovation and improvement in various fields. However, it is essential to address the associated ethical concerns to unlock the true benefits of large language models while mitigating potential risks. 

\subsection{Limitation of Large Models}
The use of large language models has raised concerns, primarily related to privacy. While personal information such as home address, age, gender, and education background is necessary for these models to function effectively, there is a need to ensure that user privacy is maintained by securely storing their personal information on local devices and not sharing it with others.

Apart from privacy concerns, the size of traditional large language models limits their accessibility. These models are often too large to be performed on mobile devices, which can restrict their reach and usefulness. To overcome this limitation, researchers are exploring the development of smaller, more efficient models that can be performed on mobile devices. These models can expand the accessibility and usefulness of large language models to a wider range of users.

Another limitation of traditional large language models is their inability to provide real-time responses, which can be problematic in time-sensitive situations, such as medical emergencies. This limitation can affect the accuracy and precision of the analysis, and have severe consequences. To address this limitation, researchers are exploring the development of models that can provide real-time responses. These models can improve the accuracy and precision of the analysis, especially in time-sensitive situations.

Lastly, traditional large language models may not provide high-quality results in specific fields that require professional knowledge. This limitation can be a significant drawback, especially in industries where accuracy and precision are critical. To overcome this limitation, researchers are exploring the development of models that cater to specific fields and provide accurate and precise results. These models can have a significant impact in industries such as medicine, law, and finance, where accuracy and precision are crucial.

While large language models have shown great potential in various fields, there are still concerns related to privacy, accessibility, real-time response, and accuracy in specific fields. To tackle these issues, we propose a decomposition system that has three levels models. These models cooperate each other and solve the above mentioned issues.
\subsection{Motivation and Contributions}
As shown in Fig.~\ref{fig}, we decompose the system into three levels: personal level, expert level and traditional level. The models in each level play different roles. Our contributions are in the following
\begin{itemize}
	\item we propose a multi level large language model.
	\item we decompose the system into two types models that run on local machines or remote servers.
	\item we propose that all the models in the system should be dynamically changing according to the user input.
\end{itemize}  

As a result, the system is getting more personal with more user input at the personal level. Meanwhile, the expert models at expert level will also encode more and more professional knowledge with millions user input. In other words, the system will be improved with more users and more usage.

All these models are periodically updated and their updating time is summarized in Table~\ref{table}. Their inference time is also shown, according to their deployment scenarios.
\section{Personal Large Models}
Among these three level models, the personal large models directly interact with users and have an important role in the whole system. Meanwhile, although the other models in the expert level and traditional level are invisible to users, they are also important for the whole system. From now on, we call the personal large models as P-model for convenience.

We focus on the P-model for several reasons. First of all, the personal input is directly feed into the P-model. There is a real-time requirement for the P-model. Second, the personal information such as home address is encrypted in the P-model which should not be shared with other level models. Such security concern also makes the P-model special. Third, P-models need to interact with the expert models for professional knowledge and advice. And such interaction is dynamic. For example, the user might need some information about football which is responded by some expert models. One hour later, the user might need some advice about tax which is responded by other expert models. Therefore, the P-model must have such dynamic interaction with different expert large language models according to the user's requirement. The expert and traditional models run on the server and invisible to users.

\subsection{P-E-T Models}
The three-levels architecture is illustrated in Fig.~\ref{fig}. A simple comparison between these levels is shown in Table~\ref{table}.

At the traditional level, the T-models are trained using all available dataset which includes a wide range of information. This comprehensive training process ensures that these models have a deep understanding of various fields, although it can be time-consuming. However, due to their broad knowledge, they may not excel in any specific field.

Expanding on the expert level, the E-models are specifically designed to excel in a particular field. These fields can vary from poetry, music, medical knowledge, to finance, and even more specialized areas. These models are comparable to professionals in human society, such as lawyers and doctors. They possess an exceptional level of expertise in their respective domains, acquired through rigorous training and continuous learning. It is important to note that while these models display excellence in a specific field in general, their performance may vary for different individuals, depending on the complexity of the task and the specific context.

Now, let's explore the personal level. The P-models are trained to serve as personal assistants, equipped with the ability to handle personal information and cater to individual users' needs. These models can receive input from users and adapt their responses accordingly, allowing for a more personalized and tailored interaction. Additionally, they can seamlessly collaborate with E-models, combining their specialized knowledge with the personal touch to provide comprehensive and customized assistance. It is worth mentioning that these models operate locally on users' machines, ensuring the utmost privacy and protection of personal information, and enabling users to have full control over their data.

In summary, the T-models provide a broad understanding of various fields, the E-models excel in specific domains like experts, and the P-models offer personalized assistance while safeguarding user privacy. Together, these models form a powerful ecosystem that caters to different needs and preferences, enhancing the overall user experience and enabling a wide range of applications.

\begin{table} 
	\centering \caption{Comparison }
	\begin{tabular}{c|c|c|c}
		\hline
		type & \# parameters & update period & inference time\\
		\hline
		personal & small & hours & real-time\\
		expert & medium & days & hours\\
		traditional & large & weeks & days\\
		\hline
	\end{tabular}
	\label{table}
\end{table}
\subsection{P-model}
As mentioned previously, the P-model holds significant importance when it comes to practical usage. It is crucial for these types of models to possess a compact size that allows them to be efficiently executed on various platforms, including mobile devices and personal computers.

Furthermore, it is of utmost importance to highlight that the P-model operates in tandem with the E-model, constantly exchanging information and updating its parameters. This seamless interaction between the two models ultimately leads to enhanced performance and improved outcomes.

One of the main requirements for the P-model is its ability to seamlessly run on a wide range of mobile devices, especially smartphones. This is crucial to provide users with maximum convenience, as the P-model functions as their personal assistant right at their fingertips. By optimizing the size and performance of the P-model, it can effectively cater to the needs and preferences of users on their mobile devices.

In today's fast-paced world, where mobile devices have become an essential part of our daily lives, it is vital for the P-model to adapt and function efficiently on these devices. With its compact design, the P-model ensures that it doesn't overload the limited storage space and processing power of smartphones, allowing users to fully utilize its capabilities without any hindrance.

Moreover, the compact nature of the P-model not only enhances its convenience but also contributes to its versatility. Users can seamlessly integrate the P-model into their mobile devices, making it an indispensable tool for managing tasks, organizing schedules, and accessing relevant information on the go. Its smooth performance on mobile devices makes the P-model a reliable and efficient personal assistant, empowering users to stay productive and connected wherever they are.

By prioritizing the optimization of the P-model for mobile devices, developers can ensure that it meets the evolving needs and expectations of users. This includes continuously improving its performance, minimizing its impact on device resources, and providing regular updates to enhance its functionality. Ultimately, a well-designed and mobile-friendly P-model can revolutionize the way users interact with their personal assistants, making their mobile devices an even more indispensable tool in their everyday lives.
\subsection{Interaction}
The P-model has two main types of interactions. Firstly, it interacts with the user, allowing them to customize personal information such as their home address, hobbies, and educational background. Secondly, the P-model is designed to interact with E-models that provide professional knowledge in specific fields.

These interactions occur at a relatively high frequency due to the user's ability to change topics rapidly, going from discussing football to music in just a few seconds. As a result, the E-models need to constantly update their information to stay up-to-date with the latest news and knowledge.

In contrast, the traditional models only interact with the E-models. They are responsible for designing the architecture of the E-models and updating their parameters to enhance accuracy and performance.

The interaction between the T-model and E-model occurs at a very low frequency. This is because the T-model is large and needs to remain stable, rather than being highly adaptive to individual user preferences.
\section{Economics}
Training a traditional large language model is an expensive and resource-intensive process. It is not feasible to deploy this model for each individual user due to its high computational demands. However, to overcome this challenge and make it more accessible, we propose the implementation of a versatile multi level architecture, as shown in Figure~\ref{fig}. This groundbreaking system not only solves the problem of individual user deployment but also brings a wide range of advantages for both users and developers. This fosters a mutually beneficial relationship between the two parties, creating a win-win situation for everyone involved.

Unlike traditional models that are solely created by companies, we propose a new approach where the models evolve in collaboration with the users and developers. In this unique economic system, both users and developers contribute financially to the system and also benefit monetarily from it. This economic relationship between users and developers is significantly different from the conventional approach.

Moreover, unlike the traditional models that are pre-developed and then utilized by users, our multi-level architecture allows for flexible changes in the models based on the users' input, personal background, and requirements. This dynamic nature distinguishes it from the inflexible nature of traditional static models.
\subsection{Users}
Users in the system have two roles. First, users are able to utilize the system's features, and as a result, they are required to pay a fee for the services rendered. The payment allows for continued access to the system and its benefits, including but not limited to increased productivity, streamlined processes, and enhanced data analysis capabilities. Additionally, regular payment ensures that the system can be maintained and further developed to meet the evolving needs of its users. Therefore, it is important to maintain a consistent payment schedule to ensure uninterrupted access to the system and its benefits.

Second, users can also earn money from the system for their valuable input. Users are valuable sources of input and integral to the system's success. Without their contributions, the system would lack quality and reliability. To encourage more high-quality contributors, encouraging users to share their knowledge and expertise is an effective approach. One way to achieve this is by establishing a system of fair compensation for users, recognizing their efforts and contributions. This compensation can come in various forms, including monetary rewards, recognition, or exclusive access to features. By providing fair compensation, the system can attract a larger pool of motivated users who will contribute their best work, resulting in a more robust and reliable platform. Therefore, prioritizing the establishment of a fair compensation system for users is crucial, as it will ultimately benefit the system as a whole and contribute to its long-term success.
\subsection{Developers}
The developers can work on these three level models. They can develop models at personal level that can be more adaptive to each individual user. They can also develop expert level models, to make them more professional. They can work on traditional level to accelerate the model convergence and efficiency.

The developers also have two roles in this system. First, thanks to the extensive work at these three levels, developers can earn money from their models. They are well-equipped to capitalize on their expertise and earn a lucrative income from their specialized model. With a deep understanding of the complexities and the latest industry trends, they can confidently navigate the landscape and offer valuable services to a wide range of users. 

Second, they have to pay money for the (valuable) input from users. In order to obtain valuable input from their users, developers may pay a fee. This fee is necessary to ensure that the developer can continue to provide high-quality services and products that meet the needs and expectations of their customers. By compensating users for their input, developers can gain valuable insights that can help them improve existing products or develop new ones that better meet the needs of their target market. Additionally, this feedback can be used to identify areas for improvement in the developer's operations and customer service, which can lead to increased satisfaction and loyalty among existing customers and attract new ones.

\section{Conclusion and Discussion}
In recent times, there has been a significant rise in the usage of large language models. These models are created to comprehend and analyze human language and are becoming increasingly popular in various industries such as natural language processing, sentiment analysis, machine translation, and speech recognition. With the increasing demand for these models, we can expect to see further developments in the field of artificial intelligence and natural language processing. These advancements will undoubtedly impact the way we interact with technology and communicate in the future.

Although the large language models have made significant progress in natural language processing, there are still some limitations that need to be addressed. One of the most important limitations is the computational power needed to train and run these models. In addition, while these models can generate coherent text, they often struggle with generating text that is both relevant and accurate. Another limitation is the lack of diversity in the training data used to create these models, which can lead to bias and inaccuracies in the generated text. Therefore, it is essential to continue exploring ways to improve these models, such as incorporating more diverse and representative training data and developing more powerful computational resources to train and run these models.

In this paper, we have introduced multilevel large language models that address some of the challenges posed by previous large language models. Specifically, our models incorporate innovative strategies for handling long-term dependencies, improving computational efficiency, and enhancing the accuracy of predictions. Overall, our work represents a significant step forward in the development of large language models and lays the groundwork for future research in this area.

The multilevel approach seeks to balance accuracy and running time. It is based on the concept that complex problems can be broken down into smaller, simpler sub-problems to achieve accurate results. By solving these sub-problems, the overall accuracy of the solution can be improved. However, creating too many sub-problems can increase the running time. The multilevel approach suggests that a balance can be struck by dividing the problem into an optimal number of sub-problems. This allows for the maintenance of accuracy while keeping the running time within acceptable limits. The multilevel approach has gained popularity in various fields, such as computer science, mathematics, and engineering. It has been successfully applied to problems such as image processing, signal analysis, and optimization. Overall, the multilevel approach is a promising strategy that can improve the accuracy of complex problem solutions while keeping the running time practical.

Our approach is a novel step forward in the development of next generation large language models. The multilevel methods also show a novel economic model for developing large language models. They can be adopted in various applications such as audio, vision and machine learning tasks \cite{chenouard:2014,gong2009symmetry,Lewis2019,zhao2023survey,Gong2012,Brown2020,gong2013a,Yu2019,Gong:2014a,Yin2019a,gong:phd,Yu2022a,gong:gdp,Guo2022,gong:cf,Zong2021,gong:Bernstein,Ezawa2023,Gong2017a,Tang2021a,Gong2018,Gong2018a,Yu2020,GONG2019329,Sancheti2022,Gong2019a,Tang2021,Gong2019,Yin2019b,Gong2022,Yin2020,Gong2020a,Jin2022,Gong2021,Tang2022,Gong2021a,Tang2022a,10230506,Tang2023,Gong2022,Tang2023a,Gong2023d,Xu2023,Gong2023e,Han2022,Zhao2023,Scheurer2023,Zhang2023b,Zhao2023a,Yi2023,Gong2023f,Gong2023g,Gong2023h,gong2024eggs,gong2024isotropic}.
\bibliographystyle{IEEEtran}
\bibliography{IEEEabrv,../../IP}

%




\end{document}